%% file: System_paper_2023.tex
\documentclass[lettersize,journal,table,xcdraw]{IEEEtran}
\usepackage{amsmath,amsfonts}
\usepackage{algorithmic}
\usepackage{algorithm}
\usepackage{array}
\usepackage[caption=false,font=normalsize,labelfont=sf,textfont=sf]{subfig}
\usepackage{textcomp}
\usepackage{stfloats}
\usepackage{url}
\usepackage{verbatim}
\usepackage{graphicx}
\usepackage{cite}
\usepackage{bm}
\usepackage{caption}
\usepackage{todonotes}
\usepackage{lipsum}
\usepackage[binary-units]{siunitx}
\usepackage{mathtools,xparse}
\usetikzlibrary{calc}
\usepackage{pgfplots}\pgfplotsset{compat=1.9}
\usepackage{graphicx}

\begin{document}
\include{defs}

\title{Field Robot for High-throughput and High-resolution 3D Plant Phenotyping}

\author{Felix Esser$^{*,1}$, Radu Alexandru Rosu$^{*,2}$, André Cornelißen$^{1}$, Lasse Klingbeil$^{1}$, Heiner Kuhlmann$^{1}$, Sven Behnke$^{2}$
        
\thanks{$^{*}$ Equal contribution.}
\thanks{$^{1}$ University of Bonn, Institute of Geodesy and Geoinformation.}
\thanks{$^{2}$ University of Bonn, Autonomous Intelligent Systems.}
}

\markboth{}
{Esser \MakeLowercase{\textit{et al.}}: Field Robot for High-throughput and
High-resolution 3D Plant Phenotyping}
\setcounter{figure}{-2}
\makeatletter
\let\@oldmaketitle\@maketitle
\renewcommand{\@maketitle}{\@oldmaketitle
    \input{imgs_tex/teaser.tex}
    }
\makeatother

\maketitle

\begin{abstract}
With the need to feed a growing world population, the efficiency of crop production is of paramount importance. To support breeding and field management, various characteristics of the plant phenotype need to be measured---a time-consuming process when performed manually. We present a robotic platform equipped with multiple laser and camera sensors for high-throughput, high-resolution in-field plant scanning. We create digital twins of the plants through 3D reconstruction. This allows the estimation of phenotypic traits such as leaf area, leaf angle, and plant height. We validate our system on a real field, where we reconstruct accurate point clouds and meshes of sugar beet, soybean, and maize.
\end{abstract}


\section{Introduction}

    With today's population growth and adverse climatic conditions, the need for crop monitoring and intervention has increased. However, manual field management is costly and laborious, which has encouraged the development of automated systems. In this article, we present such a system for high-throughput and high-resolution plant phenotyping. Phenotyping is the measurement and description of functional and structural plant traits such as fruit quality, leaf area, and biomass. These traits are critical for field management (fertilization, irrigation, or weeding) and breeding, where genotypes with desired traits are selected.

    Existing phenotyping methods are diverse and range from visual observation of the crop to cutting down plants for chemical or structural analysis. In controlled environments such as greenhouses, phenotyping facilities have been established to automate the process. 
    However, in the field, where the plant is in a natural environment subject to weather conditions and competition from neighboring plants, the phenotyping process is still mostly performed manually. Although remote sensing platforms such as UAVs, airplanes, and satellites are increasingly being used for measurements from higher altitudes, the limited resolution of their sensors and plant self-occlusion can cause issues when phenotyping at the leaf and sub-leaf scale.
    %
    The ground robot presented in this work enables automated high-throughput, high-resolution phenotyping in the field. It is equipped with multiple laser and camera sensors to capture accurate plant representations. The characteristics of the two system modalities are complementary: the 3D laser scanner can capture larger sections of the field in a short time, while the cameras focus more on individual plants and are suitable for reconstructing high-resolution color together with geometry. The acquisition system is precisely georeferenced using Real-Time Kinematic (RTK) GNSS positioning, allowing it to be combined with other relevant geospatial information, such as soil maps, and to revisit individual plants in the field.
    %
    \reffig{fig:teaser} shows the robot and exemplary 3D reconstructions from both sensing modalities. Both measurement systems provide a high geometric resolution, resolving plant organs such as individual leaves and even smaller structures. The laser-based reconstruction uses two laser-line triangulation sensors whose profile scans are georeferenced using a factor graph-based pose optimization technique. The camera-based system uses the neural implicit method of PermutoSDF~\cite{hashsdf} to reconstruct 3D geometry and texture from 2D images only.
    %
    We validate the presented phenotyping system on the \mbox{PhenoRob} experimental field. Accurate point clouds and textured meshes of sugar beet, soybean, and maize were created. 
    In addition, we compare the laser and camera systems to each other given high-quality ground truth from a static indoor laser scanner. 
    \noindent In summary, our contributions are:
    \begin{itemize}
    \item a ground robot equipped with high-resolution laser sensors, a camera dome, and a georeferencing system,
    \item methods to address outdoor lighting and localization challenges,
    \item demonstration of neural implicit representation reconstructions on a wide variety of plants imaged in the field,
    \item comparison between laser scanner and camera system against ground-truth 3D measurements, and
    \item experimental validation of the phenotyping system on a real field.
    \end{itemize}
\section{Related Work} \label{sec:related}

    \input{imgs_tex/all_sensors.tex}

    Existing field robots can be broadly categorized according to whether they are designed for intervention, such as weeding and seeding, or only for inspection and phenotyping. Although the distinction is not always clear, here we describe and compare some of the existing solutions.

    For autonomous phenotyping, Bonirob~\cite{bonirob} is a large four-wheeled robot with both electric and internal combustion engines designed for long-term scanning of sugar beet plots. It is equipped with a 4-channel multispectral JAI camera and an RGB-D Kinect v2 camera. However, the sensor does not have the accuracy and resolution required for highly detailed plant reconstruction. Later, Bonirob was also extended to support weeding tasks.
    TerraSentia\footnote{Earthsense: TerraSentia \url{https://www.earthsense.co}} is another robot designed for phenotyping but on a smaller scale. The robot is small enough to fit under the canopy and navigate between crop rows. It captures camera images to the front, sides, and top, as well as horizontal lidar distance measurements, which are analyzed offline to estimate multiple plant traits.

    In contrast, our system is designed to navigate between rows and is flexible and high-resolution enough to scan plants from less than a centimeter to almost a meter high.

    Several works exist on robots specialized in field intervention, such as Robotti2\footnote{Agrointelli: Robotti \url{https://agrointelli.com}}, Oz4403\footnote{Naio Technologies: Oz \url{https://www.naio-technologies.com/en/oz}}, Dino4\footnote{Naio Technologies: Dino \url{https://www.naio-technologies.com/en/dino}},  and BonnBot-I~\cite{ahmadi2022bonnbot}, all focusing on weed or pest control. Although they typically include laser scanners and cameras, the sensors are used to detect weeds or infected plants and not to reconstruct the full structure of the crop. They are also limited to observing smaller and younger plants, which is when weed control is most important. In contrast to these robots, we focus on high-resolution and accurate above-canopy 3D crop reconstructions under field conditions, with a quality that allows modeling and phenotyping at the plant and plant-organ level.
\section{Materials and Methods}
    \subsection{Field Robot} \label{sed:fieldrobot}
    Our field robot extends the Thorvald II platform~\cite{thorvald}, a lightweight modular base designed for agricultural purposes.
    Thorvald II features an electric 4-wheel drive that can be adjusted to different track widths and robot sizes. The onboard computer coordinates the motor wheels and can be remotely controlled using a Bluetooth-enabled Xbox controller, implemented in an ROS environment. This interface provides the potential for autonomous driving capabilities and enables easy communication and integration of the various sensor systems.
    We extend the default Thorvald base by an aluminum enclosure for the sensors of width and length  $\SI{1.5}{\metre} \times \SI{1.5}{\metre} $ as seen in~\reffig{fig:all_sensorsA}. Together with the wheels, which are attached to a suspension module, the robot has a height of about \SI{2}{\metre}.
    We add two additional computers to the platform for the laser and cameras. Both computers are equipped with multiple USB and Ethernet ports for easy connection to all the sensors and a large SSD hard drive for storing the captured data. They also have ROS interfaces for connecting to and triggering the cameras and laser. Processing of the data for 3D reconstruction is done offline, as it usually takes longer than the capturing session. 
    A foldable table holds a laptop while working in the field and motorized curtains on the front and rear openings reduce excessive sunlight inside the robot while capturing plant data. The robot's 48\,V lithium battery is transformed to the power needed by our sensor and compute systems.
    \subsection{Georeferencing System} \label{subsec:georef}
    For robot navigation and spatio-temporal mapping, we need a global position accuracy on the order of a single plant, i.e. a few centimeters. To achieve this, we equip the platform with an Inertial Navigation System from SBG Systems, which includes a dual-antenna multi-frequency GNSS receiver and an inertial measurement unit (IMU) as shown in~\reffig{fig:georref_gnss_imu}. 
    %
    \subsubsection{RTK GNSS Hardware}
    Real-time kinematic (RTK) GNSS positioning with centimeter accuracy requires a 4G internet connection streaming the necessary data from the reference station network provider SAPOS NRW. Additionally, the receiver computes the GNSS baseline between the front and rear antennas, which provides an absolute estimate of the UGV's heading and pitch. We use both the position and angle observations within the pose estimation as presented in section \ref{sebsec:pose_estim}.
    \begin{figure}[!t]
        \centering
        \includegraphics[width=1.0\linewidth]{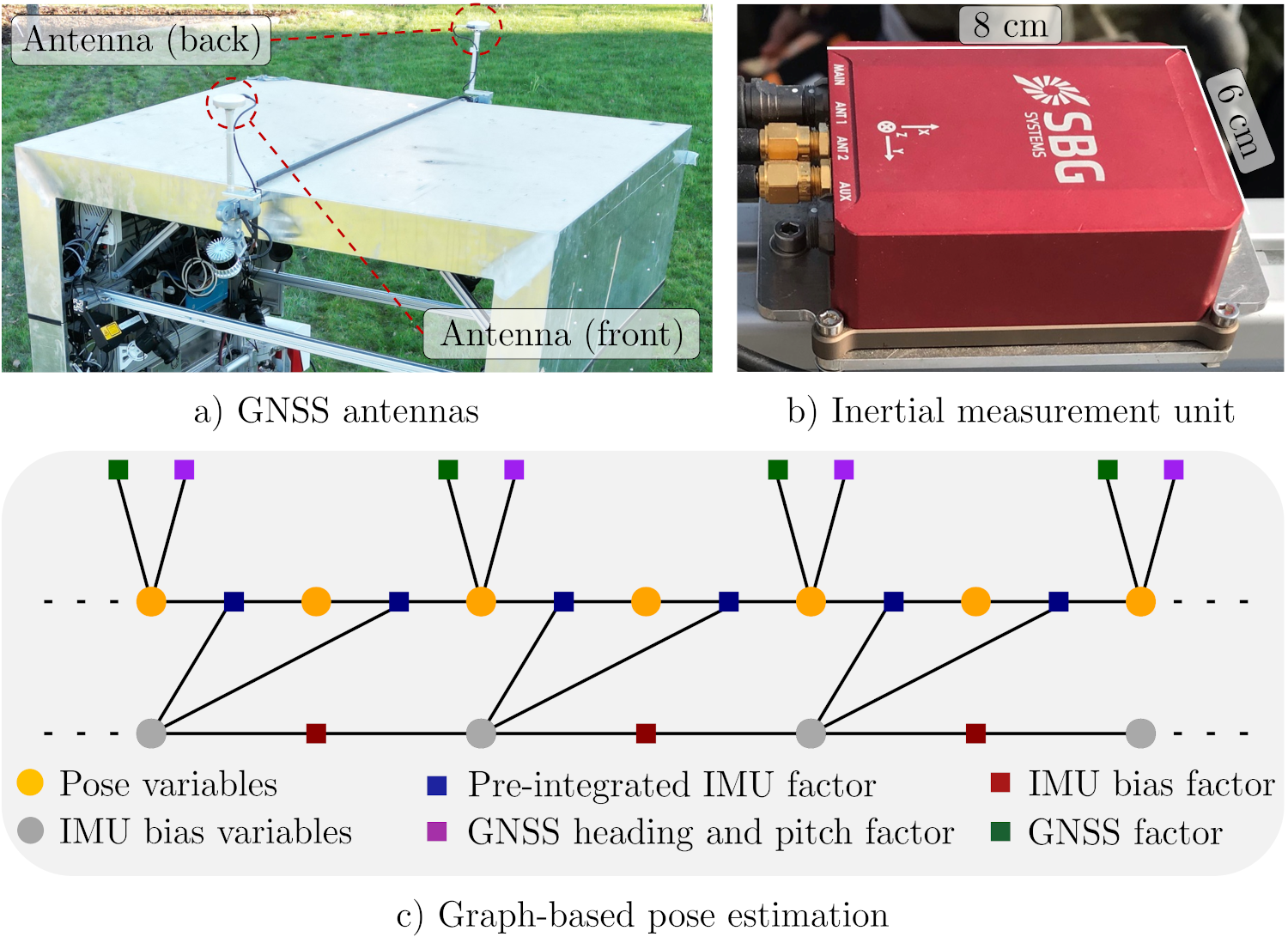}
        \caption{UGV georeferencing. a) Antennas for global positioning with centimeter accuracy using RTK GNSS. b) Inertial navigation system with integrated GNSS receiver. c) Factor-graph-based pose estimation using GNSS and IMU data.}
        \label{fig:georref_gnss_imu}
    \end{figure}
    %
    The MEMS-based industrial-grade IMU shown in \reffig{fig:georref_gnss_imu}b is installed under the robot's roof in the back. It measures in 3D accelerations and angular velocities with a sampling rate of 100\,Hz. 
    The GNSS and IMU data are streamed to an industrial-grade computer that we also use for sensor data synchronization with the laser scanning data. We use the PPS (pulse per second) signal from the GNSS receiver to assign GPS time stamps to the laser scanner measurements and link those to the estimated pose data with high precision.
    \subsubsection{Pose Estimation}\label{sebsec:pose_estim}
    To compute the poses of the robot while driving in agricultural fields, a factor-graph-based approach is used~\cite{dellaert2012factor}. In the factor graph shown in \reffig{fig:georref_gnss_imu}c, the parameters of the robot state (position and orientation) are represented as variable nodes. The factor nodes represent the non-linear functions of the sensor measurements and parameters of the variable nodes. They consist of the pre-integrated IMU factor, the GNSS factor, and the GNSS heading-and-pitch factor. The IMU factor takes into account the sensor measurements of acceleration and angular velocity by integrating them, thereby imposing constraints on the robot's change of position and orientation. The GNSS factor depends on the position measurements of the rear antenna and the heading-and-pitch factor incorporates the global orientation information of the GNSS receiver. Further variable nodes represent IMU bias parameters to compensate for its systematic errors. 
    From the linearization of the factor nodes, a normal equation system can be established to define a quadratic cost function for global pose estimation. The iSAM (incremental smoothing and mapping) algorithm~\cite{indelman2013information} is used to minimize this cost function, as implemented in the  GTSAM\footnote{GTSAM: Factor graphs for Sensor Fusion in Robotics \url{https://gtsam.org}} library. This yields smoothed robot poses to georeference the sensor data of the robot's camera and laser system for crop reconstruction.
    \subsection{Laser Scanning System}
    Typical laser scanners, often used in robotic mapping systems, do not provide the measurement precision necessary for detailed 3D reconstruction of single plants and plant organs. We therefore decided on a 3D sensing technology, which is often employed for 3D inspection tasks in factory assembly lines. Here, triangulation-based laser line scanners allow for capturing sub-millimeter accurate 3D measurements of the objects as they move through the scanner's active area. In our system, we turn this concept around by moving the scanner with the robot to scan the plants.
    \subsubsection{Laser-line Triangulation Sensors}
    The kinematic laser scanning system shown in~\reffig{fig:all_sensorsC} consists of two laser triangulation scanners (LMI Gocator 2490) attached to robot side panels at a height of \SI{1.2}{\meter} and a lateral offset of \SI{1.4}{\meter}, measuring vertically towards the ground between the wheels.
    We use two scanners to reduce occlusions due to the crops themselves. The scanners are tilted at an angle of about \SI{50}{^\circ} with respect to the side panels to capture lower plant structures.
    The sensors use a red laser (\SI{600}{\nano\metre}) diode to project a laser line into the plot (see \reffig{fig:all_sensors}c right), which is reflected to the sensor's 2D CCD array. Based on the known geometry between the center of the projector and the lens focusing on the CCD, the distance for each point along the line is triangulated, yielding 2D points in the sensor coordinate system. For more information on the methodology of laser triangulation sensors, we refer to \cite{ebrahim20153d}.
    The measurement distance ranges from 390 to 2000\,mm with a repeatability of \SI{12}{\micro\meter}. Each laser line consists of \num{1920} points, resulting in point-to-point distances along the laser line of about \SI{0.5}{\milli\meter} at a depth of \SI{1}{\meter}. Important sensor settings are the exposure time $\Delta t$ and the scan rate $r$. The exposure time needs to be adjusted to obtain proper reflections from the crop surfaces without over-exposure. The scan rate, together with the driving speed, determines the line distances along the driving direction.
    An exposure time of \SI{1200}{\micro\second} and a scan rate of 200\,Hz show reliable scans of crop structures in terms of spatial resolution and reflectivity. At a robot speed of \SI{10}{\cm / \second} and a scan rate of 200\,Hz the line spacing is 0.5\,mm.
    \subsubsection{Kinematic Laser Scanning} \label{subsubsec:direct_georef}
\begin{figure}[!t]
        \centering
        \includegraphics[width=1.0\linewidth]{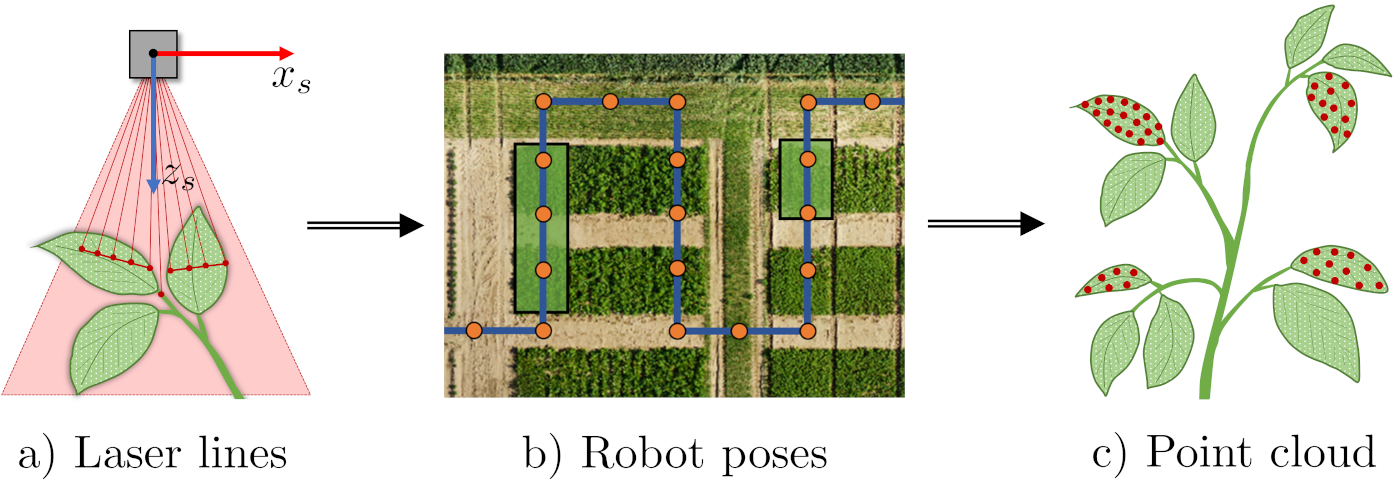}
        \caption{Kinematic laser scanning pipeline for 3D point cloud creation. a) 2D laser lines given in the sensor frame. b) Robot poses known from factor graph optimization. c) Generated 3D point cloud. See eq.~\ref{eq:direct georef}. }
        \label{fig:direct_georef}
    \end{figure}

    For 3D crop point cloud creation, we georeference the 2D laser line measurements from the scanners using the globally optimized robot poses, estimated by the georeferencing system.
    This kinematic laser scanning pipeline is shown in \reffig{fig:direct_georef} with the corresponding equation (\ref{eq:direct georef}). The line distance measurements of the laser triangulation scanner are given in the sensor coordinate frame ($s$), only populating the XZ-plane (\reffig{fig:direct_georef}a). In the first step, the measurements are transformed into the body frame ($b$) of the robot, which coincides with the IMU sensor frame, by applying the system calibration parameters. These parameters describe the orientation $\mathbf R_s^b$ and translation $[\Delta x, \Delta y, \Delta z]$ between the scanners and the body frame. They have been estimated using a plane-based calibration approach, as described in \cite{heinz2020design}. 
    \begin{equation}
        \begin{bmatrix} x_g \\ y_g \\ z_g \end{bmatrix} = \begin{bmatrix}p_x \\p_y \\p_z\end{bmatrix}_g+\mathbf{R}_b^g \left[\begin{bmatrix}
                            \Delta x \\
                            \Delta y \\
                            \Delta z
                        \end{bmatrix} +
                        \mathbf R_s^b \begin{bmatrix}
                            x_s \\
                            0 \\
                            z_s
                        \end{bmatrix} \right]
         \label{eq:direct georef}
    \end{equation}
    The robot poses (\reffig{fig:direct_georef}b), describe the position $[p_x, p_y, p_z]$ and orientation $\mathbf R_b^g$ of the robot with respect to the global coordinate system $g$. Since the poses are needed at the exact times of the laser scan measurements, we accurately tag the time of the laser scans in the same time frame as the poses and then performed a cubic interpolation of the poses with respect to the laser time. The application of this direct georeferencing procedure to the left and the right scanner data leads to the creation of a globally consistent point cloud, as seen in \reffig{fig:direct_georef}c. For more details on the georeferencing of 2D laser scanner measurements, we refer to \cite{esser2023}.
    \subsection{Color Camera System}
    To capture individual plants in the center of plots from all sides, the UGV has been equipped with a dome of high-resolution color cameras.
    \subsubsection{Camera Hardware}
    We use 20 Nikon Z7 cameras arranged around the center of the UGV, as shown in \reffig{fig:all_sensorsB}. We use the variable focal length Nikkor Z 24-70mm lens and capture images at a resolution of 8256$\times$5504\,pixels.
    The zoom of each camera is manually adjusted depending on the growth stage of the plant to maximize the effective resolution on the plant, which is in the range of 10-30\,pixels/mm.
    The cameras are attached to the robot frame with adjustable mounts for fast on-field modifications. We distribute the cameras on a hemisphere of approximately 1.5\,m radius. We also experimented with placing the cameras in stereo pairs with short baselines but found that a roughly uniform distribution works better for plant reconstruction as it minimizes possible occlusions or under-constrained areas. Diffuse lighting is provided by multiple LED panels mounted in the robot's enclosure.
    \subsubsection{Camera Triggering}
    All cameras are connected to an onboard PC via USB. We interface with the cameras using the libgphoto2 library\footnote{H. U. Niedermann and H. Figui\`ere: GPhoto2 \url{http://www.gphoto.org}} which allows to trigger all cameras synchronously. Various parameters such as aperture, ISO, and shutter speed can be controlled. This interface is made available to ROS. Before triggering the cameras, we send a signal to focus all cameras on a central patch of the image where we expect the plant of interest to be. After focusing, we send a capture signal to all cameras in parallel such that the images captured are as time-synchronized as possible. In our experiments, we observed that the maximum delay between camera captures is below \SI{33}{\milli\second} which is precise enough for our use case, as the UGV is stationary during image capture.  
    We store both JPEG and RAW formats to allow for flexible post-processing.
    \subsubsection{Camera Auto-exposure and Calibration}
    %
    During in-field capture, we want to ensure that all cameras have the same ISO, aperture, and shutter speed to obtain consistent exposures. However, it is impossible to use fixed parameters because the lighting conditions in the field are constantly changing due to the weather and passing clouds. To address this issue, we developed a method for multi-camera auto-exposure. At a frequency of 5\,Hz, all cameras capture a low-resolution 800$\times$600 image which is transferred to the onboard PC. Using all 20 images, we compute a gray-scale histogram of eight bins. We are interested in reducing the number of pixels that saturate to either too dark values (gathered in the first bin of the histogram) or too bright values (last bin), thus ensuring that the image is correctly exposed with most of the values falling within the measurable range of the camera. For this, we start with initial values for ISO, aperture, and shutter speed and also define reasonable minimum and maximum values for them. We follow a simple but robust algorithm to modify the camera parameters:
    \begin{enumerate}
        \item While the first and last bin of the histogram are not sufficiently equal, we increase or decrease the ISO of all cameras by 100 such that their values are brought closer together.  
        \item If we reach the acceptable ISO limit, we switch to modifying the aperture of all cameras by adjusting the F-stop. 
        \item If we reach the acceptable F-stop limit, we are left with modifying the shutter speed in increments of 5\,ms until the image becomes correctly exposed.
    \end{enumerate}
    We choose this priority list based on the expected effect of each parameter on the final image. Increasing the ISO can cause noise grain in the images. However, this can be alleviated since we are fusing information from multiple cameras. Therefore changing the ISO will have the highest priority since it's the easiest effect to correct. Increasing the aperture of the camera can result in a shallow depth of field and a blurred background. However, since the plant in the center of the image is expected to stay mostly sharp, the final reconstruction is usually unaffected by the depth of field. Lastly, increasing the shutter speed can cause severe motion-blurring artifacts which are difficult to correct. Therefore, we place this parameter last in our list of priorities and increase it only if the other two parameters fail to achieve a correct exposure.
    We initialize all cameras with an ISO of 400, an f/14 aperture, and \SI{10}{\milli\second} shutter speed and let our algorithm dynamically modify them based on the weather conditions.
    To obtain the camera poses, it is not sufficient to rely on offline calibration methods as the robot is non-rigid and deforms while driving, causing the relative camera poses to change. To address this issue, we estimate the camera poses using Colmap~\cite{schoenberger2016sfm} after each multi-view image capture. We initialize the camera poses with a reasonable prior for the location and orientation and perform bundle adjustment to estimate both intrinsic and extrinsic parameters.
    \subsubsection{Image Capture GUI}
    During robot operation in the field, it is often necessary to inspect the captured images and to manually trigger the cameras. For this, we use the Foxglove GUI\footnote{Foxglove Technologies Inc.: Foxglove \url{https://foxglove.dev}}. Foxglove enables us to remotely connect to the onboard PC to visualize selected ROS topics. We show in real time the captured low-resolution 800$\times$600 images from all cameras. Foxglove can be used either as a standalone app or as a web application which extends its usage to environments that don't have a ROS installation. 
    \subsubsection{3D Plant Reconstruction using Neural Implicit Surfaces}
    For the 3D reconstruction of plants given multiple 2D RGB images from different viewpoints, we take inspiration from recent neural implicit models~\cite{neus, volsdf, unisurf, hashsdf}. This family of models represents a 3D scene with a neural network that encodes both color and geometry. While the color is modeled as a function of position and viewing direction, geometry is modeled as either a signed-distance function (SDF) or an occupancy field. The neural networks are trained using volumetric rendering so that the renderings match the captured images as closely as possible. 
    We rely on neural implicit models for reconstructing plants due to their ability to recover very fine detail and complete geometry even in occluded areas. We choose the recent PermutoSDF~\cite{hashsdf} model which optimizes an SDF and a color field using local features embedded in a permutohedral lattice. \reffig{fig:hashsdf_overview} gives an overview of the PermutoSDF pipeline. We refer to Rosu and Behnke~\cite{hashsdf} for more details.
    While PermutoSDF obtains accurate models, visualizing them requires performing volumetric rendering or sphere tracing. This is both expensive and requires specialized software solutions. To address this issue, we extract a textured mesh from the trained PermutoSDF model that can be visualized in any 3D package. For our visualizations, we use EasyPBR~\cite{easypbr}.
    To obtain the mesh, we first run Marching Cubes~\cite{marching} on the reconstructed signed-distance field. Secondly, we decimate the mesh to \SI{300}{\kilo{}} faces using QSlim~\cite{qslim} to reduce triangle density without sacrificing geometric quality. We rely on texture mapping to obtain high-resolution color. For this, the mesh is UV-unwrapped using the automatic unwrapping tool from Blender\footnote{Blender Organization: Blender \url{https://www.blender.org}} which establishes a UV-parameterization. We use a 4000$\times$4000 RGB texture on the mesh and set each texel to the average color of all cameras that have an unobstructed view of the texel:
    \begin{equation}
        \mathbf{x}_i = \sum_{j=1}^{20} v_i^j \cdot \mathbf{I}_j( P_j( \mathbf{u}_i)  )
    \end{equation}
    where $\mathbf{x}_i$ is the value of $i$-th texel, $P$ projects a texel position $\mathbf{u}_i$ into the $j$-th camera coordinate system, $\mathbf{I}_j$ represents the $j$-th camera image, and $v_i^j$ is a binary visibility indicator that is $1$ if texel $i$ is visible in camera $j$ and $0$ otherwise.
    Reconstructing plants using PermutoSDF is done offline on a computer with an RTX 3090 GPU and takes approximately 30\,minutes per plant, followed by 10\,minutes for mesh extraction and texture creation. We note that PermutoSDF can also run on less powerful GPUs by slightly downsampling the images. 
    \input{imgs_tex/hashsdf_overview.tex}
    \input{imgs_tex/recon_growth.tex}

\section{Datasets and Experiments} \label{sec:exp}
We performed several experiments to validate the proposed plant phenotyping methods. 
They include field experiments with both laser-based and camera-based captures of plots and individual plants, respectively. We extract phenotypic traits from the 3D reconstructions and compare the precision and completeness of the 3D reconstructions from the cameras and the laser scanners.
    \subsection{In-field Image Capture and Laser Scanning} \label{subsec:field_capture}
    Field experiments were performed on the PhenoRob Central Experiment field at Campus Klein-Altendorf in Germany which features various crops and species that allow for a wide variety of phenotypic traits to be captured. We focused on measuring sugar beet, maize, and soybean and acquired laser and image data with our robot platform for a total of 14 days spanning from May 13 until August 2nd, 2022. This covers the bulk of the vegetation period, from the initial sprouting of the plants until the plants became too large to fit inside the robot. With the multi-camera system, we captured images from approximately 30 plants from each species. This resulted in a dataset of $\approx \SI{2.3}{\tera\byte}$ consisting of \num{24046} images. With the laser system, we captured a large dataset on August 2nd, 2022 including scans of multiple crops rows of sugar beet, soybean, and wheat leading to $\approx \SI{100}{\giga\byte} $ high-resolution point cloud data. 
    In terms of the time taken to measure and reconstruct 3D using the laser system, we need about 30\,s to collect data from one plot (1.5\,m $\times$ 3\,m) and a few minutes to post-process and create the point cloud on a standard office computer at a driving speed of 10\,cm / s and a scanner rate of 200\,Hz. For the camera system, we stop on top of each plot to trigger the cameras which takes about 2\,s followed by a copy to disk which is done in the background. We then reconstruct the scene using the 20 images on a computer with an RTX 3090 which takes about 30\,min.
    In addition to the images and laser scan data, we also recorded the position and orientation of the robot's georeferencing system which enables us to embed our reconstructions within other datasets that were captured in the same field, as shown in~\reffig{fig:gps_zoomin} and~\reffig{fig:soy_wheat_sugarbeet}.
    %
    \subsection{3D Reconstructions of Individual Plants and Plots}  \label{subsec:field_pheno}
    \label{subsec:field_phenotyping}
\begin{figure*}[!ht]
        \includegraphics[width=1.0\linewidth]{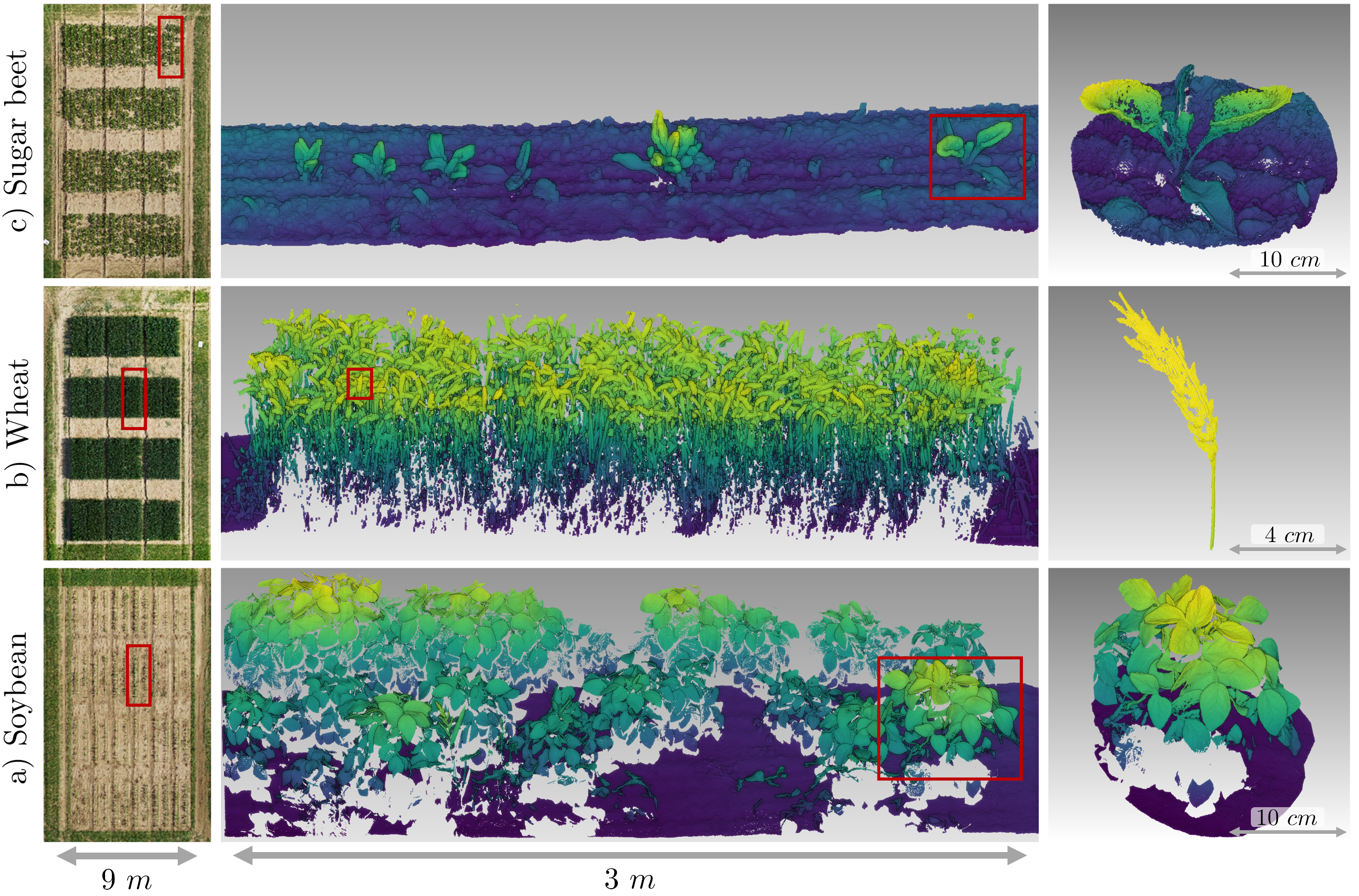}
        \caption{Point clouds created with the laser scanning system of the robot while driving in the field. a)-c) Soybean, wheat, and sugar beet in the UAV image of the field (left), on plot level (middle), and single plant level (right).}
        \label{fig:soy_wheat_sugarbeet}
    \end{figure*}

    \reffig{fig:gps_zoomin} and \reffig{fig:soy_wheat_sugarbeet} show camera-based and laser-based 3D reconstructions of plants in the field, respectively. Both types of reconstructions are overlaid with UAV images, highlighting the highly accurate georeferencing. The camera-based reconstructions provide detailed textured meshes of the measured individual plants. The output of the laser scanning system is a high-resolution 3D point cloud of entire plots. 
    In the soybean and sugar beet plots, single leaves are visible. Within the wheat plot, even the ears and the stems are recognizable.
    \input{imgs_tex/gps_zoomin.tex}
    \reffig{fig:recon_growth} shows image-based reconstructions of the same soybean plant over multiple days and its estimated height. Analyzing plant growth is needed for growth models and allows to identify potential plant stressors that can stunt growth. Therefore, 4D (3D space + time) reconstruction is of great interest to breeders and plant scientists. 
    Structural analysis and tracking of plant organs such as~\cite{chebrolu2021registration} registers the plant temporally and provides insight into how the structure of the plant changes as it grows.
    \subsection{Precision and Completeness} \label{subsec:prec}
\begin{figure}[!t]
        \centering
        \includegraphics[width=1.0\columnwidth]{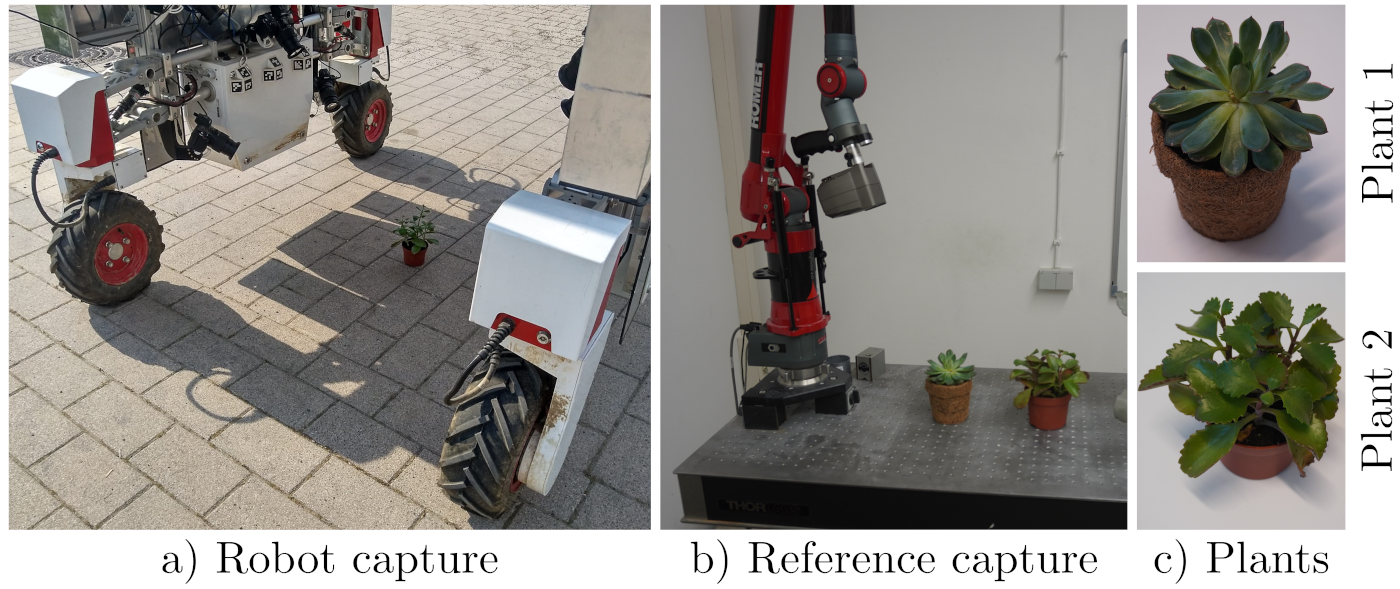}
        \caption{Measurements for precision and completeness analysis. a) Plants 1 and 2 are captured with the robot's camera and laser system. b) Reference capture using a precise scanning system in the lab. c) Plant~1: Hen-and-chick (Echeveria Pelusida), Plant~2: Lamb's tail (Chiastophyllum oppositifolium).}
        \label{fig:measurements_ref}
    \end{figure}

\begin{figure*}[!ht]
        \centering
        \includegraphics[width=1.0\linewidth]{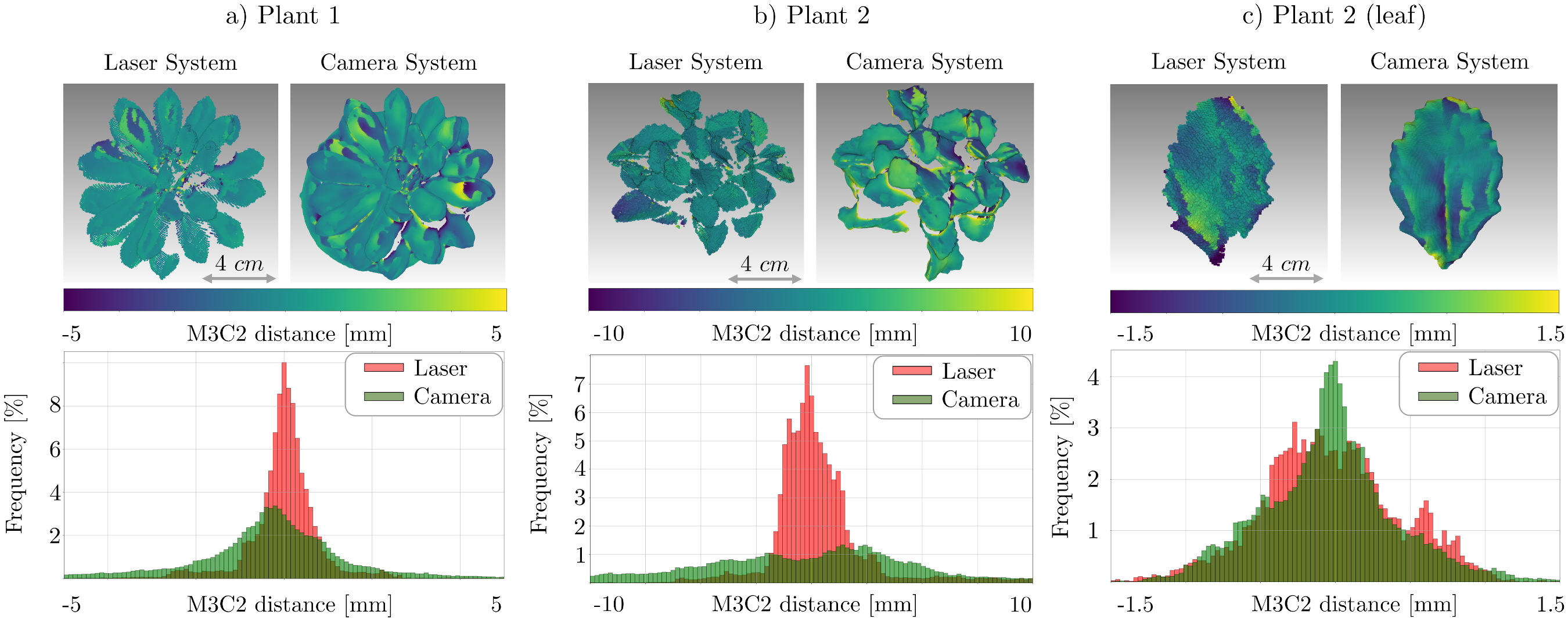}
        \caption{ M3C2 comparison between the reference measuring arm scan and the 3D reconstructions of the camera and laser system for the two plants of \reffig{fig:measurements_ref}a. a), b) M3C2 distances and their histograms of Plant~1 and Plant~2 for laser and camera system. c) M3C2 distances and histograms for an example leaf of plant 2.}
        \label{fig:m3c2_p123}
    \end{figure*}

    To analyze the reconstruction quality of the two approaches in terms of precision and completeness, we generate 3D reference point clouds of two distinct plants using a high-precision scanning system in a lab. We capture and scan the same plants using cameras and laser sensors of the robot system in an outdoor environment, mimicking the in-field situation, and compare the reconstruction results with the lab reference.  
    \reffig{fig:measurements_ref}b shows the lab scanner. It consists of a measuring arm and a laser triangulation sensor commonly used in industrial applications. The sensor poses are calculated from joint encoder measurements by forward kinematics and used to reference the laser measurements to create a high-resolution 3D reference point cloud with a point accuracy below 1\,mm. This measurement system was used to create the popular Pheno4D dataset~\cite{Pheno4D}.
    For this experiment, we chose two semi-rigid plants to minimize plant movements while scanning and between the scans enabling a proper comparison of the reconstructions. \reffig{fig:measurements_ref}c shows the chosen Hen-and-chick (Echeveria Pelusida) and the Lamb's tail (Chiastophyllum oppositifolium) plants. 
    We obtain 3D textured meshes from the captured images and 3D point clouds from the laser measurements. We call the standard deviation of the distances between the reconstructed mesh or point cloud and the reference point cloud 'precision' and provide it as a measure of the reconstruction quality. In the absence of georeferencing information about the reference cloud, we first register the reconstruction results with the reference using the ICP (iterative closest point) algorithm. We then calculate the distances using the multi-scale model-to-model (M3C2) algorithm~\cite{lague2013accurate}. M3C2 uses the surface normal vectors of the reference point cloud to build point correspondences at averaged core points and computes signed distances between the point clouds. For detailed information on the M3C2 method, see~\cite{lague2013accurate}. For plant 2, we additionally extract leaves and perform an ICP fine registration for each of the five leaves individually. The standard deviation of the M3C2 distances is used as a value for the reconstruction precision of the robot's two 3D capturing systems.
    To evaluate the completeness of the reconstruction, we compare the leaf area of the reconstruction with the ground truth. To obtain the ground truth leaf area, we use the reference point cloud and perform surface reconstruction using the ball-pivoting algorithm~\cite{bernardini1999ball} which generates a 3D triangle mesh. The areas of all triangles are summed up to estimate the total leaf area. Since the output of the robot's laser scanning system is also a point cloud, we repeat this pipeline for these scans as well. We compute the differences in leaf area for the laser and camera systems relative to the ground truth.
    \reffig{fig:m3c2_p123} shows the distribution of the M3C2 distances of the laser- and the camera-based reconstruction for the two plants and one example leaf of the Lamb's tail plant. We observe that the reconstruction from the camera system exhibits a long tail distribution compared to the laser which is more peaked around zero and therefore more precise. The precision of the single leaf reconstruction on the right-hand side of \reffig{fig:m3c2_p123} is in general better due to the separate leaf-level fine registration before the comparison, since the sugar beet leaf comparison includes an additional fine ICP registration the deviations from the reference are smaller. 
    \input{table.tex}
    \reftab{table:m3c2_leafarea_cmp} reports the reconstruction precision and the leaf area differences to the reference scans for five example leaves of plant 2. The precision of the laser system ranges from 0.17\,mm (Leaf~5) up to 0.48\,mm (Leaf 1) with a mean of 0.27\,mm. In comparison, the precision of the image-based reconstruction ranges from 0.24\,mm (Leaf~3) to 0.55\,mm (Leaf~1) and a mean of 0.43\,mm.
    The comparison w.r.t. leaf areas are shown in the right part of~\reftab{table:m3c2_leafarea_cmp}. The leaf areas estimated from the laser scans deviate from the reference from -14.9\% (Leaf~5) to 1.73\% (Leaf~3) with an overall absolute mean difference of the leaves of 6.5\%. The camera system shows leaf area differences between -3.6\% (Leaf~3) and 7.8\% (Leaf~2) and an absolute average difference to the reference of 1.9\%. We observe that the laser system tends to underestimate the leaf area more than the camera system. This is mostly because the point cloud from the laser scanning system can contain holes caused by self-occlusion by the plant itself. In contrast, the camera system uses more viewing angles for the reconstruction, resulting in more complete plant structures.  
    The advantage of the camera system is that it provides a textured mesh with a high-resolution color texture, whereas the laser provides only 3D information and laser intensity. The results show that the laser reconstructions are more accurate but less complete than the camera system. However, the camera system struggles to recover detail in reflective areas such as raindrops on the plant while the laser has problems with bright sunlight conditions and wind-induced plant movement during the scans. The two systems therefore complement each other and have different advantages and limitations.
\section{Summary and Outlook} \label{sec:summary}
We presented a ground robot for in-field phenotyping. We developed an integrated system equipped with laser scanners, cameras, and accurate pose estimation capable of reconstructing 3D models of crops in the field with high throughput, resolution, and precision. Using the reconstructed plants over multiple days of field acquisition, we show that it is possible to obtain georeferenced 3D models that offer enough resolution to reconstruct single plants and plant organs to derive phenotypic traits such as leaf area, leaf angles, and plant height.  
We believe that our system can help automate various phenotyping tasks that currently require significant manual work in the field. In the future, we plan to extend our methods by recovering structural plant models, performing automated point cloud segmentation for crop organ detection, and registering plants over time for a better understanding of the growth processes.
\section*{Acknowledgement}
This work was funded by the Deutsche Forschungsgemeinschaft (DFG, German Research Foundation) under Germany’s Excellence Strategy–EXC 2070–390732324.
\bibliographystyle{IEEEtran}
\bibliography{refs}{}
\end{document}

%% file: defs.tex
\newcommand{\ALEX}[1]{{\color{orange}{\bf ALEX: #1}}}
\newcommand{\FELIX}[1]{{\color{blue}{\bf FELIX: #1}}}
\newcommand{\TODO}[1]{{\color{orange}{\bf #1}}}
\newcommand{\NEW}[1]{{\color{blue}{#1}}}

\newcommand{\IGNORE}[1]{}

\definecolor{ph-purple}{RGB}{129, 39, 232}
\definecolor{ph-blue}{RGB}{5, 131, 227}
\definecolor{ph-blue-v2}{RGB}{35, 211, 255}
\definecolor{ph-blue-v4}{RGB}{16, 33, 75}
\definecolor{ph-gray}{rgb}{0.5, 0.5, 0.5}
\definecolor{ph-orange}{RGB}{227, 127, 5}
\definecolor{ph-green}{RGB}{0, 135, 124}
\definecolor{ph-yellow}{RGB}{235, 201, 52}
\definecolor{ph-light-green}{RGB}{181, 209, 21}
\definecolor{ph-red}{RGB}{250, 101, 60}
\definecolor{ph-gold}{RGB}{255, 223, 0}
\definecolor{ph-metallic-gold}{RGB}{212, 175, 55}
\definecolor{ph-silver}{RGB}{192, 192, 192}
\colorlet{g}{ph-gold!90!black}
\colorlet{s}{ph-silver!60}
\colorlet{ph-orange-light}{ph-orange!70}
\colorlet{ph-blue-light}{ph-blue!70}
\colorlet{ph-purple-light}{ph-purple!70}
\colorlet{ph-green-light}{ph-green!70}
\definecolor{ph-light-gray}{rgb}{0.75, 0.75, 0.75}

\newcommand{\reffig}[1]{Fig.~\ref{#1}}
\newcommand{\reftab}[1]{Tab.~\ref{#1}}
\newcommand{\refsec}[1]{Sec.~\ref{#1}}
\newcommand{\refequ}[1]{Eq.~\ref{#1}}

\DeclarePairedDelimiter{\abs}{\lvert}{\rvert}
\DeclarePairedDelimiter{\norm}{\lVert}{\rVert}


\newcommand{\OneBlob}{\text{ob}}
\newcommand{\PosEnc}{\text{freq}}
\newcommand{\Spherical}{\text{sph}}
\newcommand{\Identity}{\text{id}}
\newcommand{\StopGradient}{\text{sg}}

\newcommand{\cv}{g}
\newcommand{\CV}{G}
\newcommand{\selectionProb}{c}
\newcommand{\acv}{\hat{\cv}}
\newcommand{\aCV}{\hat{\CV}}
\newcommand{\cvCoef}{\alpha}
\newcommand{\cvShape}{\bar{\cv}}
\newcommand{\Params}{\theta}
\newcommand{\cvParams}{\Params_{\cv}}
\newcommand{\CVParams}{\Params_{\CV}}
\newcommand{\acvParams}{\Params_{\acv}}
\newcommand{\aCVParams}{\Params_{\aCV}}
\newcommand{\pdfParams}{\Params_{\PdfMC}}
\newcommand{\selectionProbParams}{\Params_{\selectionProb}}
\newcommand{\nisParams}{\Params_{\NIS}}
\newcommand{\cvShapeParams}{\Params_{\cvShape}}
\newcommand{\cvCoefParams}{\Params_{\cvCoef}}
\newcommand{\NIS}{\mathrm{NIS}}
\newcommand{\conditionals}{y}
\newcommand{\auxConditionals}{\kappa}

\newcommand{\JacobianProduct}{J}

\newcommand{\GenericParams}{\phi}

\newcommand{\GradParams}{\nabla_{\!\Params}}
\newcommand{\GradCV}{\nabla_{\!\cvParams}}

\newcommand{\cvc}{\alpha}
\newcommand{\cvh}{h}
\newcommand{\CVH}{H}

\newcommand{\numAuxDims}{E}
\newcommand{\auxDims}{\mathbf{\xi}}
\newcommand{\enc}{\mathrm{enc}}
\newcommand{\encOut}{\mathbf{y}}
\newcommand{\interpWeight}{\mathbf{w}}
\newcommand{\nn}{m}
\newcommand{\hash}{h}
\newcommand{\primeNumber}{\pi}
\newcommand{\smoothstep}{S_1}

\newcommand{\entriesPerLevel}{T}
\newcommand{\entry}{t}
\newcommand{\featuresPerEntry}{F}
\newcommand{\feature}{f}
\newcommand{\levels}{L}
\newcommand{\level}{l}
\newcommand{\resolution}{N}
\newcommand{\minResolution}{N_\mathrm{min}}
\newcommand{\maxResolution}{N_\mathrm{max}}
\newcommand{\perLevelScale}{b}

\newcommand{\BigO}{\mathcal{O}}

\newcommand{\pos}{\mathbf{x}}
\newcommand{\posy}{\mathbf{y}}
\newcommand{\dir}{\omega}
\newcommand{\diro}{\omega}
\newcommand{\diri}{\omega_\mathrm{i}}
\newcommand{\diriRV}{\Omega}
\newcommand{\normal}{\mathbf{n}}

\newcommand{\Real}{\mathbb{R}}
\newcommand{\R}{\mathbb{R}}
\newcommand{\Z}{\mathbb{Z}}
\newcommand{\Scal}{\mathcal{S}}

\newcommand{\LatticeParams}{\theta}
\newcommand{\NetParams}{\Phi}
\newcommand{\colorval}{\mathbf{c}}
\newcommand{\ViewDir}{\mathbf{v}}
\newcommand{\CamOrig}{\mathbf{o}}
\newcommand{\Pixel}{\mathbf{p}}
\newcommand{\sdf}{\mathbf{s}}
\newcommand{\GeomFeat}{\bm{\chi}}
\newcommand{\SigmoidSlope}{h}
\newcommand{\Loss}{\mathcal{L}}
\newcommand{\tangent}{\bm{\eta}}
\newcommand{\RandomVec}{\bm{\tau}}

\newcommand{\lipc}{k}
\newcommand{\liploss}{\mathcal{L}_\text{Lipschitz}}
\newcommand{\softplus}{\mathrm{softplus}\,}

\newcommand{\vecFont}[1]{\mathsf{#1}}

%% file: imgs_tex/teaser.tex
\begin{tikzpicture}

\def\imgteaser{./imgs/teaser/plant_teaser.png} 
\let\WTT\relax
\let\HTT\relax
\newlength{\WTT}
\newlength{\HTT}
\settowidth{\WTT}{\includegraphics{\imgteaser}}
\settoheight{\HTT}{\includegraphics{\imgteaser}}
\newcommand\ShiftXLarge{4.4}
\newcommand\ShiftYText{-2.5}
     
\node[inner sep=0pt] (d0) at (\ShiftXLarge*0,0)
   {\includegraphics[ width=.25\textwidth
   	]{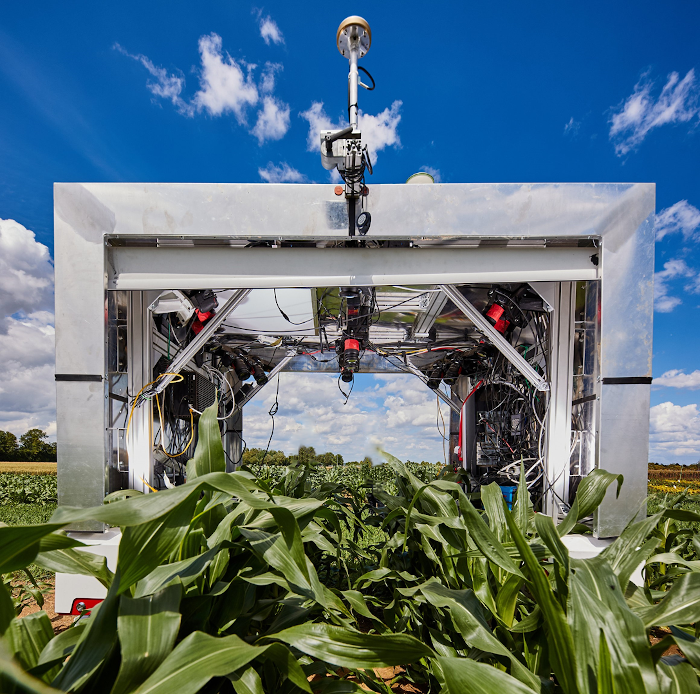}};

\node[inner sep=0pt] (d1) at (\ShiftXLarge*1,0)
   {\includegraphics[trim=.25\WTT{} .0\HTT{} .25\WTT{} .15\HTT{},clip,  width=.25\textwidth
   	]{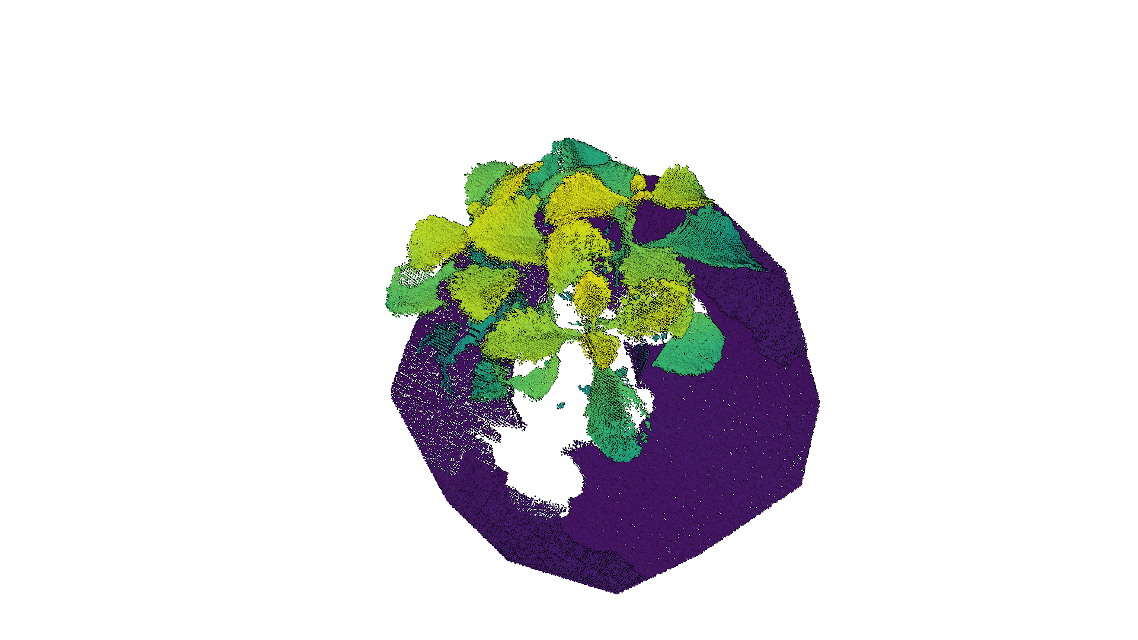}};

\node[inner sep=0pt] (d3) at (\ShiftXLarge*2,0)
   {\includegraphics[trim=.25\WTT{} .0\HTT{} .25\WTT{} .15\HTT{},clip, width=.25\textwidth
    ]{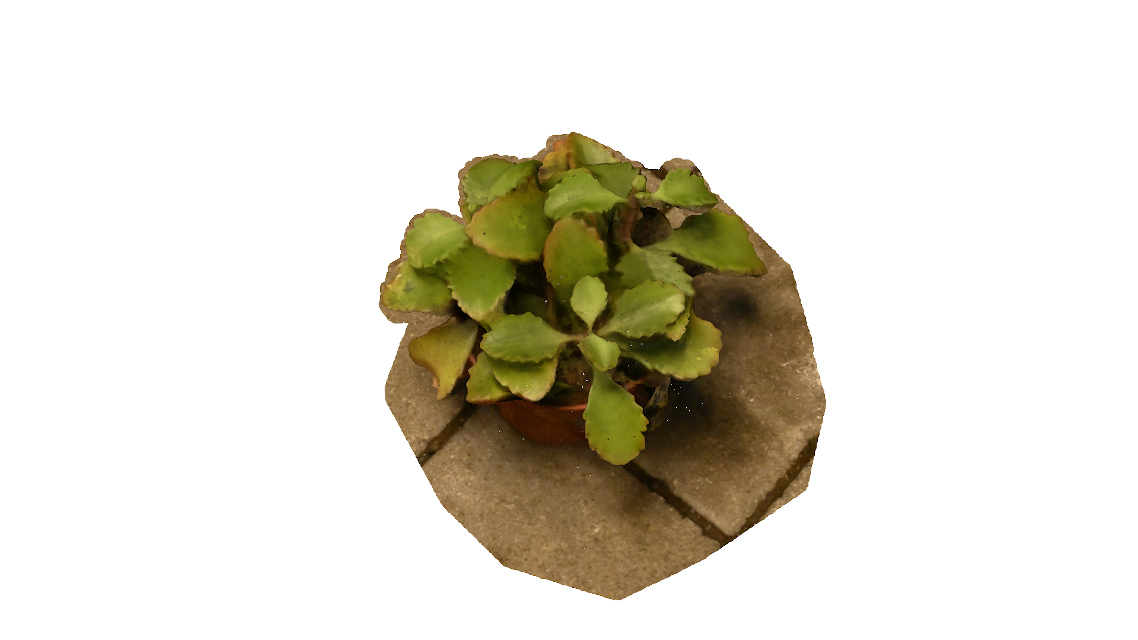}};

\node[inner sep=0pt] (d3) at (\ShiftXLarge*3,0)
   {\includegraphics[trim=.25\WTT{} .0\HTT{} .25\WTT{} .15\HTT{},clip, width=.25\textwidth
   	]{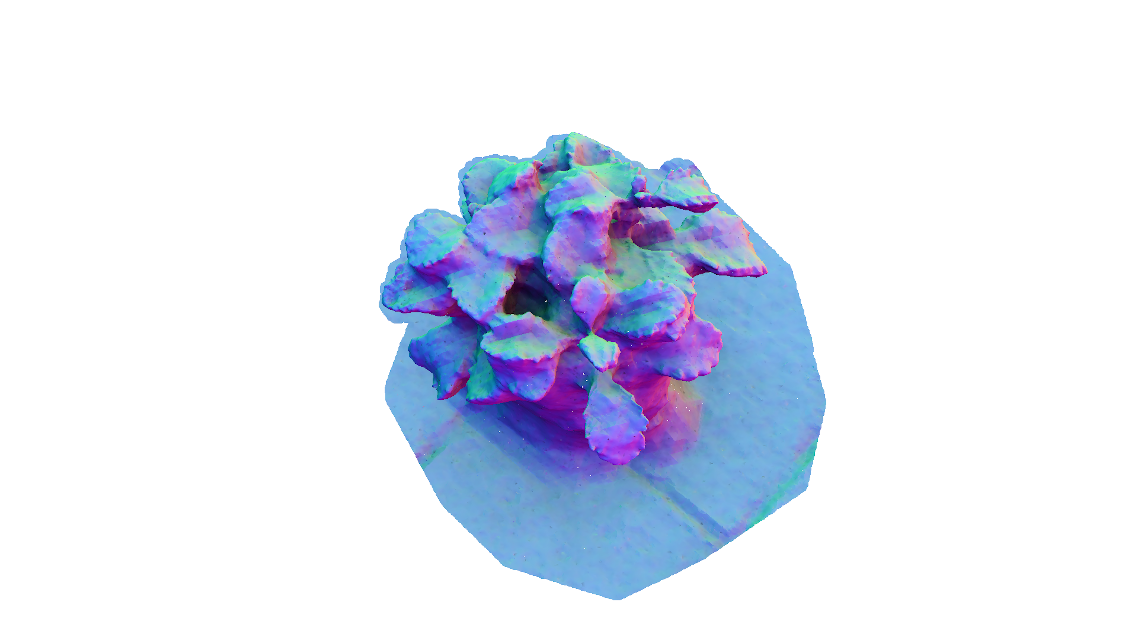}};

\node at (0,\ShiftYText) {\footnotesize a) Ground robot with multimodal sensors};
\node at (\ShiftXLarge*1+\ShiftXLarge*0.1, \ShiftYText) {\footnotesize b) Laser-based reconstruction};
\node at (\ShiftXLarge*2+\ShiftXLarge*0.5, \ShiftYText) {\footnotesize c) Textured mesh reconstruction (color + geometry)};

\end{tikzpicture}

\captionof{figure}{ a) UGV for high-resolution, georeferenced 3D in-field crop reconstruction with RTK, laser scanners and camera dome. b) Point cloud created with the laser scanning system. c) Textured mesh reconstructed from the multi-camera system.}
\label{fig:teaser}

%% file: imgs_tex/all_sensors.tex
\begin{figure*}
\centering
\small\sffamily

\subfloat{ \label{fig:all_sensorsA} }
\subfloat{ \label{fig:all_sensorsB} }
\subfloat{ \label{fig:all_sensorsC} }

\begin{tikzpicture}

\def\imgalls{./imgs/all_sensors/DSC07687_s_enchanced.JPG} 
\newlength{\WALLS}
\newlength{\HALLS}
\settowidth{\WALLS}{\includegraphics{\imgalls}}
\settoheight{\HALLS}{\includegraphics{\imgalls}}

\newcommand\ShiftXLarge{3.5} 

\node[inner sep=0pt] (r) at (\ShiftXLarge*0,0)
    {\includegraphics[trim=.1\WALLS{} .05\HALLS{} .05\WALLS{} .4\HALLS{},clip, height=4cm,
    	]{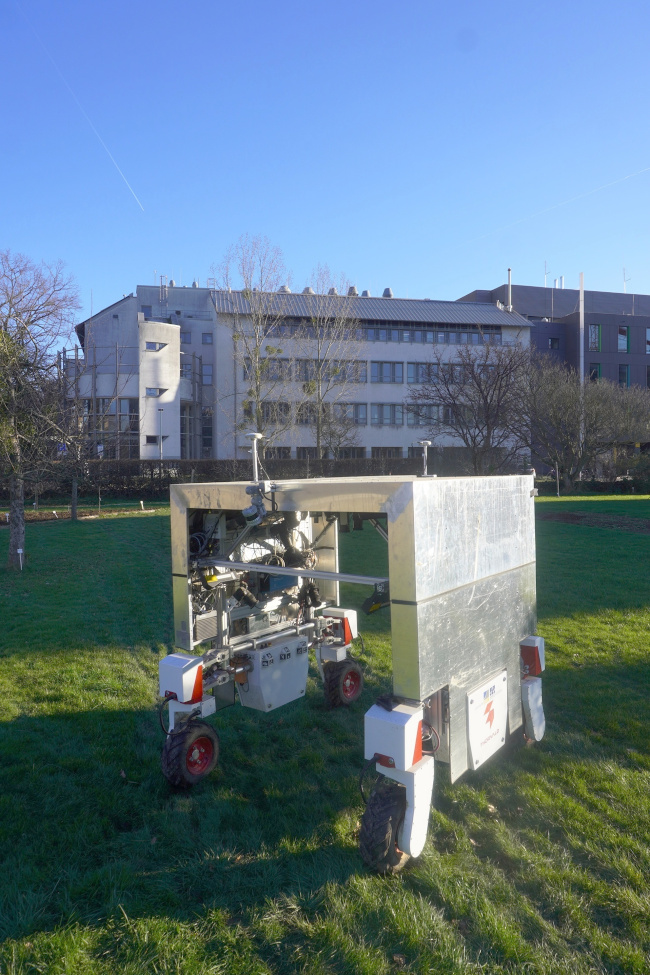}};
\node[inner sep=0pt] (c0) at (\ShiftXLarge*1,0)
    {\includegraphics[trim=.0\WALLS{} .0\HALLS{} .0\WALLS{} .0\HALLS{},clip, height=4cm
    	]{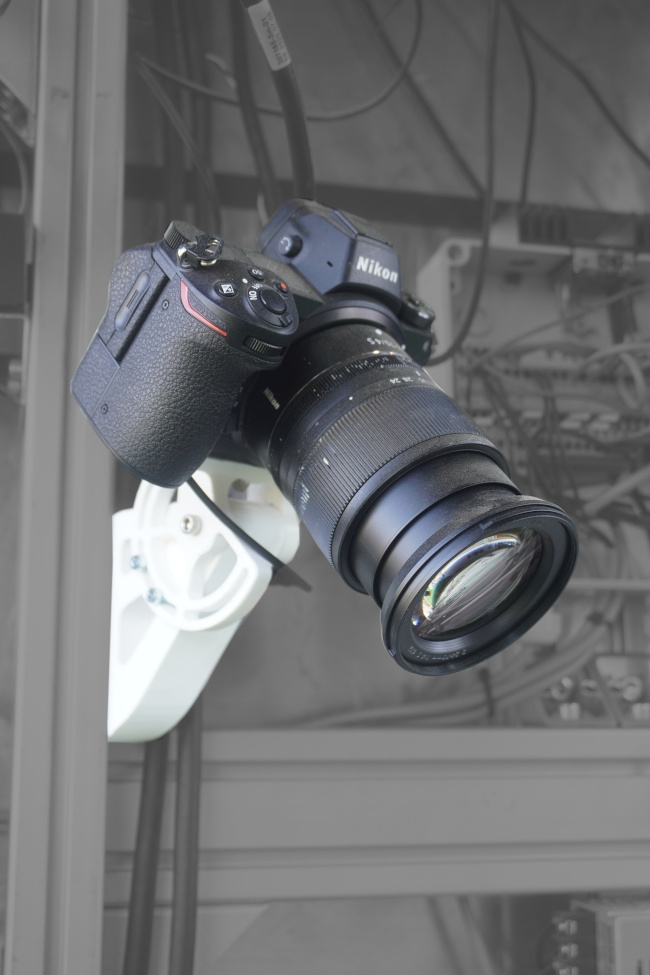}};
\node[inner sep=0pt] (c1) at (\ShiftXLarge*2,0)
    {\includegraphics[trim=.0\WALLS{} .0\HALLS{} .0\WALLS{} .0\HALLS{},clip, height=4cm
        ]{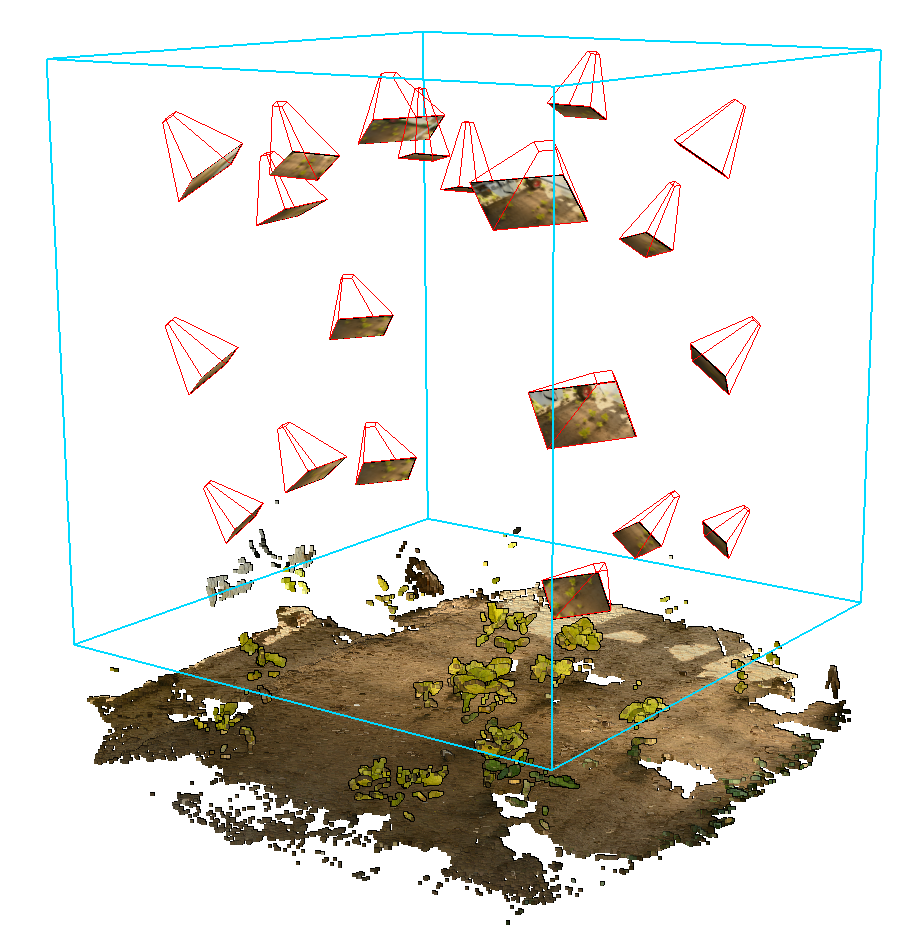}};

\node[inner sep=0pt] (l0) at (\ShiftXLarge*3,0)
    {\includegraphics[ height=4cm
    	]{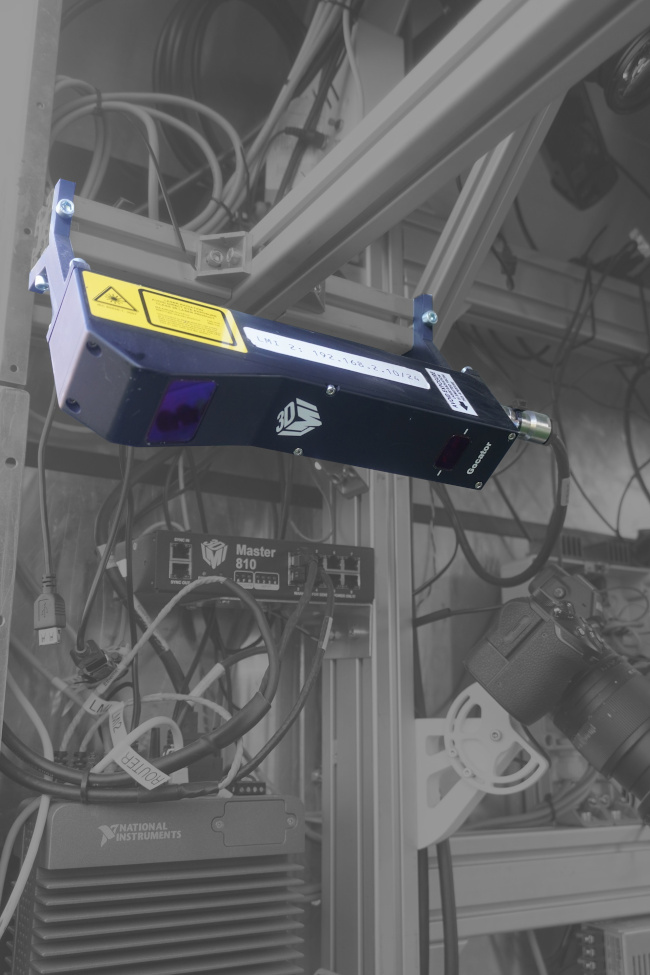}};
     \node[inner sep=0pt] (l1) at (\ShiftXLarge*4,0)
    {\includegraphics[ height=4cm
    	]{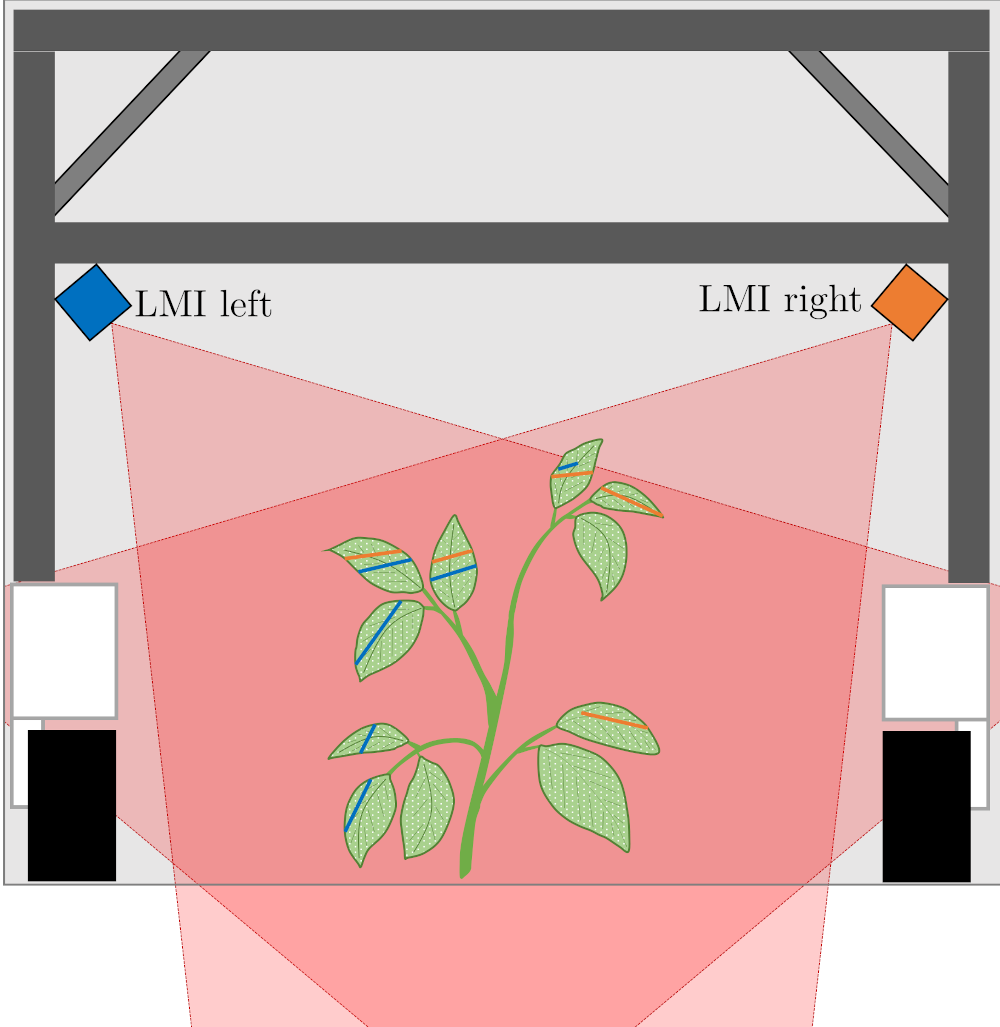}};
     
\draw[] (c0.north west) rectangle (c1.south east) {};
\draw[] (l0.north west) rectangle (l1.south east) {};

\node at ( $ (c0.north west) + (0.4,-0.3) $) {  \textcolor{white}{$\bm{\times}$\bf20}  };
\node at ( $ (l0.north west) + (0.4,-0.3) $) {  \textcolor{white}{$\bm{\times}$\bf2}  };

\node at (0.0, -2.3) {\footnotesize a) Ground robot };
\node at (5.5, -2.3) {\footnotesize b) Camera system };
\node at (12.4, -2.3) {\footnotesize c) Laser system };

 \draw[-latex,ph-blue, line width=0.3mm] (4.15, 0.6) arc
[
start angle=150,
end angle=30,
x radius=0.9cm,
y radius =0.75cm
] ;
 \draw[-latex,ph-blue, line width=0.3mm] (10.5, 1.0) arc
[
start angle=150,
end angle=30,
x radius=1.0cm,
y radius =0.75cm
] ;

\end{tikzpicture}

\caption{ \textbf{a)} UGV equipped with a multimodal sensor system for in-field plant phenotyping. \textbf{b)} 20 Nikon Z7 cameras are placed in a dome-like manner around the UGV's center. \textbf{c)} Two profile laser scanners capture entire plots while driving at low speed. } \label{fig:all_sensors}
\end{figure*}

%% file: imgs_tex/hashsdf_overview.tex
\begin{figure}
\centering
\small\sffamily
\begin{tikzpicture}

\node[inner sep=0pt] (d0) at (0,0)
    {\includegraphics[ width=1.0\columnwidth
    	]{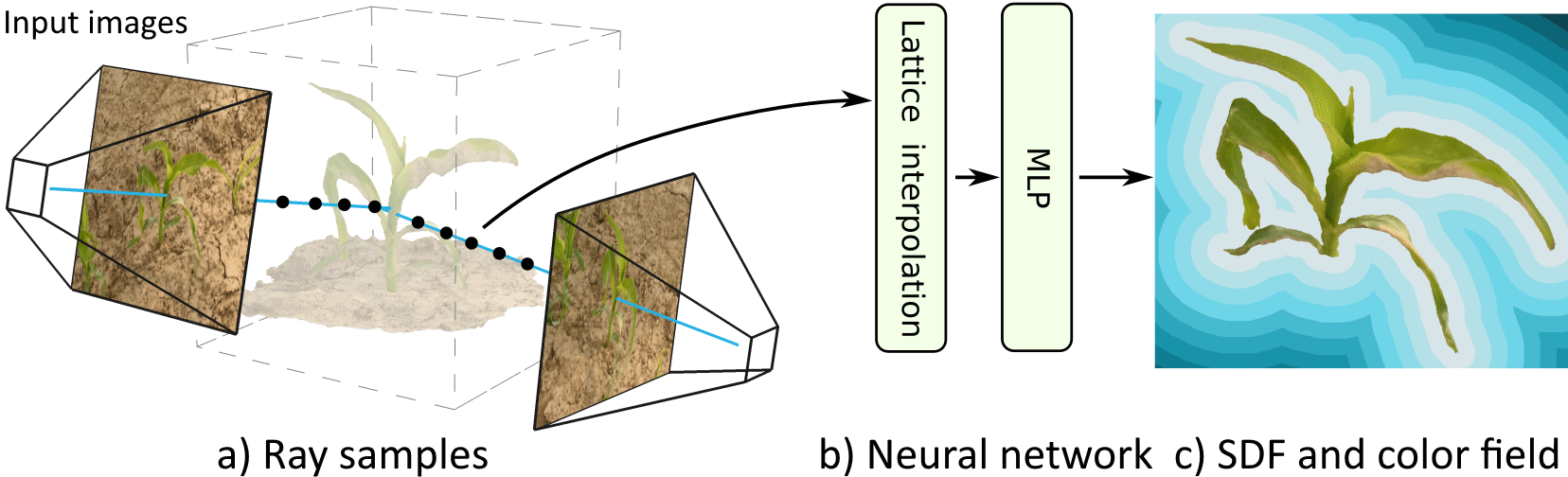}};

\end{tikzpicture}
\caption{  3D plant reconstruction from multiple 2D RGB images through volumetric rendering. We follow the approach of PermutoSDF~\cite{hashsdf} and shoot rays through every pixel of the 20 cameras (a). 3D point samples on the ray are passed through a neural network (b) to output both geometries as a signed-distance function (SDF) and a color field (c). The network is trained so that the resulting rendered images match the given captured images.} 
\label{fig:hashsdf_overview}
\end{figure}

%% file: imgs_tex/recon_growth.tex
\begin{figure}
\centering
\small\sffamily
\begin{tikzpicture}

\def\imgGrow{./imgs/recon_growth/bean/rgb/05_30_rgb_310_clean.png} 
\newlength{\WGR}
\newlength{\HGR}
\settowidth{\WGR}{\includegraphics{\imgGrow}}
\settoheight{\HGR}{\includegraphics{\imgGrow}}

\newcommand\ShiftXLarge{2.8} 
\newcommand\ShiftY{-2.55}


\node[inner sep=0pt] (d0) at (\ShiftXLarge*0,0)
    {\includegraphics[ width=.33\columnwidth
    	]{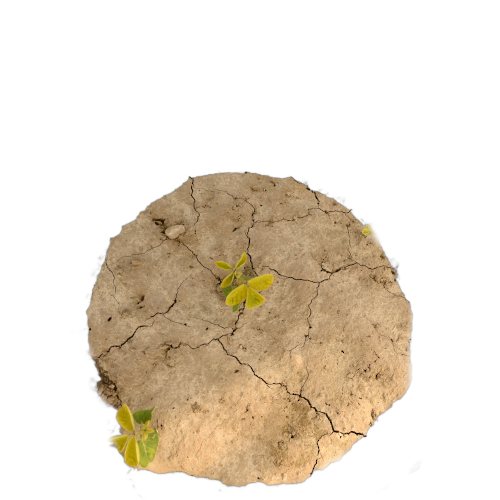}};
\node[inner sep=0pt] (d1) at (\ShiftXLarge*1,0)
    {\includegraphics[trim=.1\WGR{} .0\HGR{} .1\WGR{} .0\HGR{},clip, width=.33\columnwidth
    	]{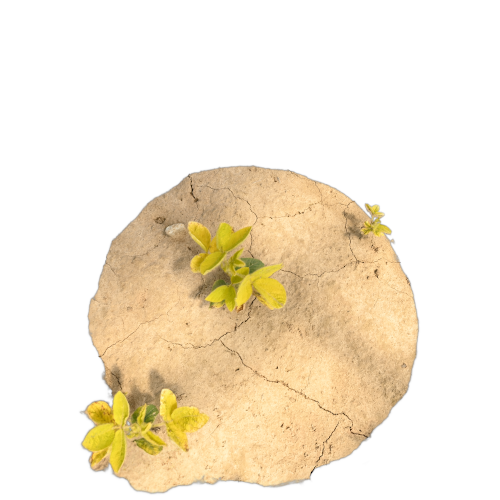}};
\node[inner sep=0pt] (d3) at (\ShiftXLarge*2,0)
    {\includegraphics[trim=.1\WGR{} .0\HGR{} .1\WGR{} .0\HGR{},clip, width=.33\columnwidth
    	]{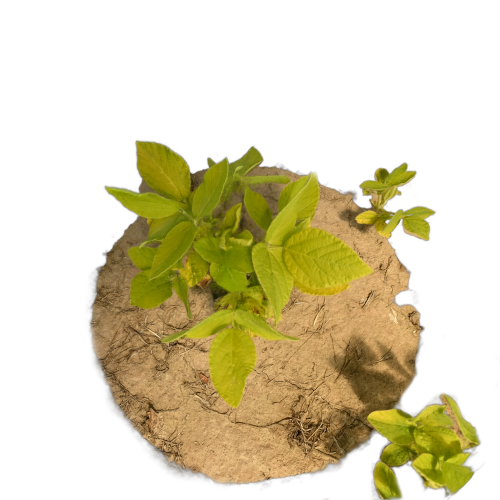}};

\node[inner sep=0pt] (d0) at (\ShiftXLarge*0, \ShiftY)
    {\includegraphics[trim=.1\WGR{} .0\HGR{} .1\WGR{} .0\HGR{},clip, width=.33\columnwidth
    	]{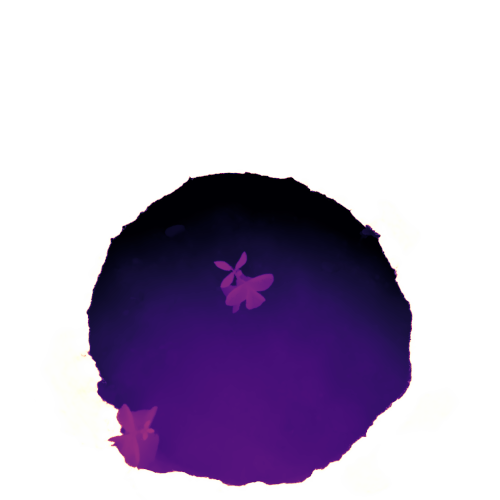}};
\node[inner sep=0pt] (d1) at (\ShiftXLarge*1, \ShiftY)
    {\includegraphics[trim=.1\WGR{} .0\HGR{} .1\WGR{} .0\HGR{},clip, width=.33\columnwidth
    	]{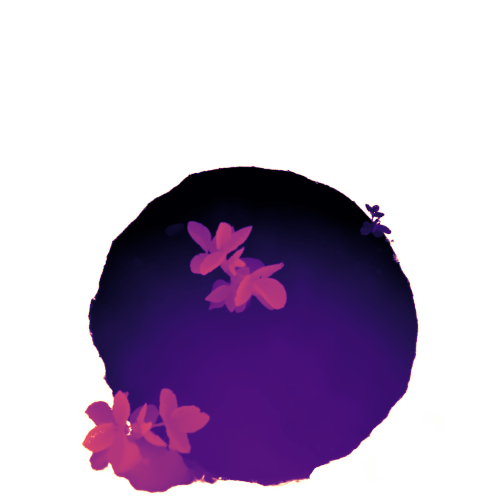}};
\node[inner sep=0pt] (d3) at (\ShiftXLarge*2, \ShiftY)
    {\includegraphics[trim=.1\WGR{} .0\HGR{} .1\WGR{} .0\HGR{},clip, width=.33\columnwidth
    	]{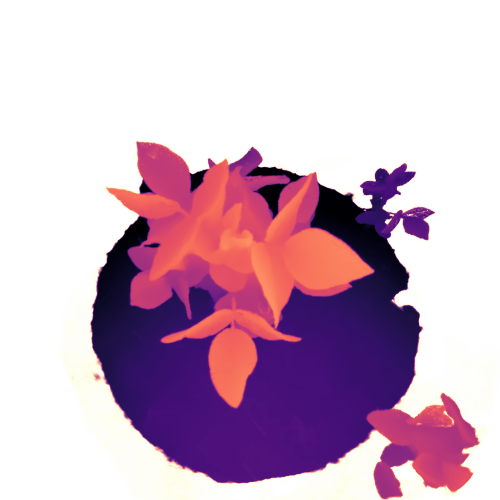}};

\begin{scope}[x=0pt,y=0cm,shift={(-0.34cm,1.2cm)}]
    \begin{axis}[ 
      width=0.96\linewidth,
      height=4.0cm,
      line width=0.5,
      grid=major,
      grid style={dashed},
      ylabel={\small Plant height(cm)},
      y tick label style={},
      xticklabels={,,},
      ybar interval,
    ]

  \addplot[black, fill = ph-red] coordinates
      {(0, 3.6) (1,4.2) (2,3.7) (3,4.0) (4,4.8) (5,4.9) (6,6.2) (7,7.2) (8,9.5) (9,11.4)};

    \end{axis}
\end{scope}
     
\draw [-stealth](\ShiftXLarge*0.5, -4.6) -- (\ShiftXLarge*1.5, -4.6);
\node at (\ShiftXLarge*1,-4.8) {\footnotesize Time};

\node [rotate=90] at (-1.5, -0.5) {\footnotesize Predicted RGB};
\node [rotate=90] at (-1.5, -3.2) {\footnotesize Predicted depth};

\draw [-stealth, dashed](0.55, 1.1) -- (0.0, 0.6);
\draw [-stealth, dashed](3.05, 1.1) -- (2.9, 0.6);
\draw [-stealth, dashed](5.7, 1.1) -- (5.65, 0.8);

\end{tikzpicture}
\caption{Image-based 3D reconstruction over multiple days. The same plant is captured over the course of nine days and textured meshes are reconstructed for each individual day. The computed plant height is shown over time and the predicted RGB and depth from the PermutoSDF network are shown for three days.} \label{fig:recon_growth}
\end{figure}

%% file: imgs_tex/gps_zoomin.tex
\begin{figure*}[!ht]
\centering

    \small\sffamily
    \begin{tikzpicture}
    
    \newcommand\Square[1]{+(-#1,-#1) rectangle +(#1,#1)}
    
    \newcommand\ShiftXLarge{3.0} 
         
    \node[inner sep=0pt] (d0) at (\ShiftXLarge*0,0)
        {\includegraphics[ width=.16\textwidth
        	]{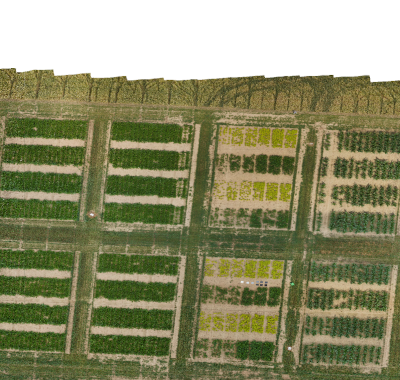}};
    
    \node[inner sep=0pt] (d1) at (\ShiftXLarge*1,0)
        {\includegraphics[ width=.16\textwidth
        	]{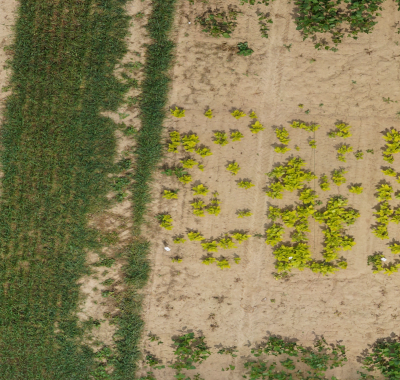}};
    
    \node[inner sep=0pt] (d2) at (\ShiftXLarge*2,0)
        {\includegraphics[ width=.16\textwidth
        	]{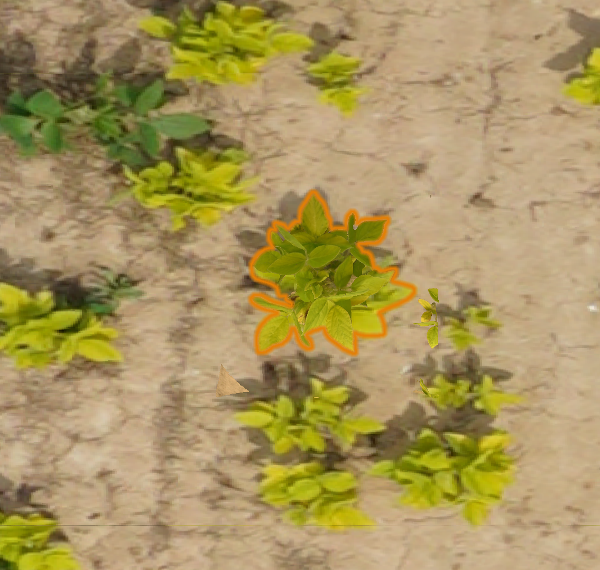}};
    
    \node[inner sep=0pt] (d3) at (\ShiftXLarge*3,0)
        {\includegraphics[ width=.16\textwidth
        	]{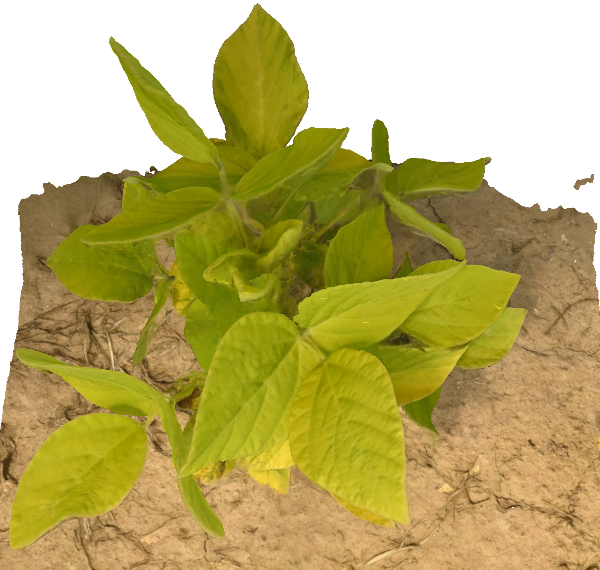}};
    
    \node[inner sep=0pt] (d4) at (\ShiftXLarge*4 - 0.1 ,0)
        {\includegraphics[ width=.151\textwidth
        	]{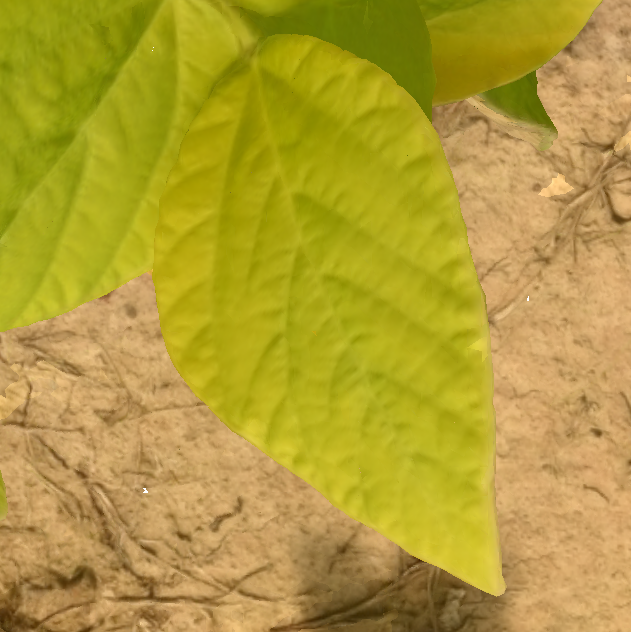}};
    
    \node[inner sep=0pt] (d5) at (\ShiftXLarge*5 - 0.3,0)
        {\includegraphics[ width=.151\textwidth
        	]{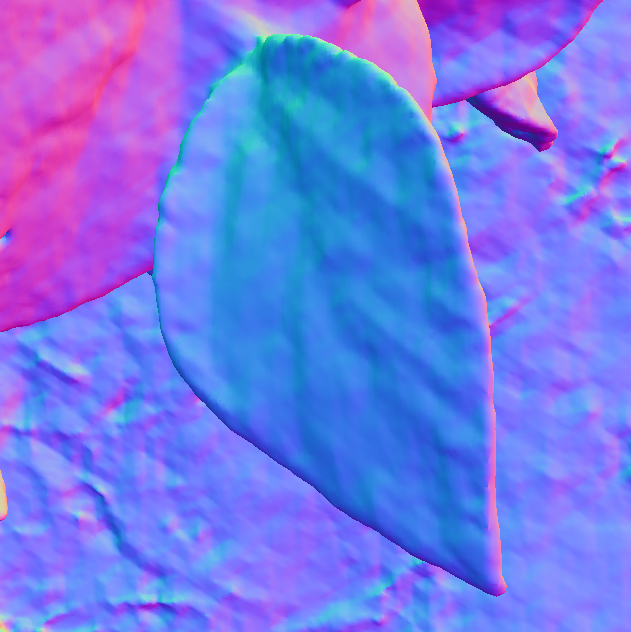}};
    
    \node[dashed,ph-blue, ultra thick] at ($(d0.center) + (0.13, 0.0)$) [minimum size=0.4cm,draw] (s0) {};
    \draw[dashed,ph-blue, ultra thick] (s0.north east) -- (d1.north west);
    \draw[dashed,ph-blue, ultra thick] (s0.south east) -- (d1.south west);
    \node[dashed,ph-blue, ultra thick] at ($(d1.center) + (0.0, 0.0)$) [minimum size=0.55cm,draw] (s1) {};
    \draw[dashed,ph-blue, ultra thick] (s1.north east) -- (d2.north west);
    \draw[dashed,ph-blue, ultra thick] (s1.south east) -- (d2.south west);
    \node[dashed,ph-blue, ultra thick] at ($(d2.center) + (0.1, 0.1)$) [minimum size=0.8cm,draw] (s2) {};
    \draw[dashed,ph-blue, ultra thick] (s2.north east) -- (d3.north west);
    \draw[dashed,ph-blue, ultra thick] (s2.south east) -- (d3.south west);
    \node[dashed,ph-blue, ultra thick] at ($(d3.center) + (0.2, -0.7)$) [minimum size=0.88cm,draw] (s3) {};
    \draw[dashed,ph-blue, ultra thick] (s3.north east) -- (d4.north west);
    \draw[dashed,ph-blue, ultra thick] (s3.south east) -- (d4.south west);
    
    \draw [ph-blue](-1.4, -1.6) -- (4.5, -1.6); 
    \draw [ph-orange] (4.5, -1.6) -- (7.5, -1.6);
    \draw [ph-green] (7.5, -1.6) -- (16.0, -1.6);
    
    \node at (2.0,-1.9) {\footnotesize UAV image};
    \node at (6.0,-1.9) {\footnotesize UAV$+$\textcolor{ph-orange}{mesh}};
    \node at (11.0,-1.9) {\footnotesize Textured mesh};
    
    \end{tikzpicture}

     \vspace*{-1mm}
    \caption{Georeferencing of the 3D textured meshes reconstructed from multi-camera images. Our field robot globally localizes with RTK while capturing images. This allows us to embed our highly detailed single-plant reconstructions into an ortho mosaic map captured by a UAV.}  
    \label{fig:gps_zoomin}
\end{figure*}

%% file: table.tex
\begin{table}[h!]
\centering \small
\caption{ Mean M3C2 distances and leaf areas compared to the reference scans for the Lamb ’s tail plant. The means of the leaf area comparisons in the last row are average absolute percent differences.}
\setlength{\tabcolsep}{5pt}
\renewcommand{\arraystretch}{1.1}
\begin{tabular}{c|cc|ccc}

                             & \multicolumn{2}{c|}{$\sigma_{M3C2}$ {[}mm{]}}                                                                                             & \multicolumn{3}{c}{Differences of Leaf Area }                                                        \\ \hline
\multicolumn{1}{c|}{Leaf} & \multicolumn{1}{c}{Laser} & \multicolumn{1}{c|}{Camera}  & \multicolumn{1}{l}{Laser [\%]} & \multicolumn{1}{l}{Camera [\%]} & \multicolumn{1}{l}{Ref. [$cm^2$]} \\  \hline 
1    & 0.48     &  \multicolumn{1}{c|}{0.55}  &   -3.3\%       & 4.1\%       & 23.78      \\ \hline
2   & 0.18     &  \multicolumn{1}{c|}{0.33}  &   ~-6.2\%       &7.8\%       & ~8.65       \\ \hline
3   & 0.32     &  \multicolumn{1}{c|}{0.35}  &  1.73\%       & ~-3.6\%       &~9.38       \\ \hline
4 & 0.22     &  \multicolumn{1}{c|}{0.67}  &  ~-6.3\%       & ~-1.2\%       & 13.18       \\ \hline
5 & 0.17     &  \multicolumn{1}{c|}{0.24}  &   -14.9\%       &~-2.4\%      & ~8.54        \\ \hline
mean & 0.27     &  \multicolumn{1}{c|}{0.43}  & 6.5\%           & ~1.9\%        &   \\    
\end{tabular}
\label{table:m3c2_leafarea_cmp}
\end{table}